%% file: main.tex
\documentclass[10pt,twocolumn,letterpaper]{article}
\usepackage{algorithm}
\usepackage{algorithmic}
\usepackage{cvpr}              
\input{preamble}
\definecolor{cvprblue}{rgb}{0.21,0.49,0.74}
\usepackage[pagebackref,breaklinks,colorlinks,allcolors=cvprblue]{hyperref}

\def\paperID{} 
\def\confName{CVPR}
\def\confYear{2026}

\title{QD-PCQA: Quality-Aware Domain Adaptation for Point Cloud Quality Assessment}

\author{
Guohua Zhang\textsuperscript{1}
\quad
Jian Jin\textsuperscript{2}
\quad
Meiqin Liu\textsuperscript{1}\thanks{Corresponding author: mqliu@bjtu.edu.cn}
\quad
Chao Yao\textsuperscript{3}
\quad
Weisi Lin\textsuperscript{2}
\\
\textsuperscript{1}Beijing Jiaotong University 
\quad
\textsuperscript{2}Nanyang Technological University
\\
\textsuperscript{3}University of Science and Technology Beijing
\\
{\tt\small 24125207@bjtu.edu.cn},
{\tt\small jianj008@gmail.com},
{\tt\small mqliu@bjtu.edu.cn},
\\
{\tt\small yaochao1986@gmail.com},
{\tt\small wslin@ntu.edu.sg}
}
\usepackage{multirow}
\usepackage{float}  
\usepackage{booktabs}
\begin{document}
\maketitle
\input{sec/0_abstract}  

\input{sec/1_intro}
\input{sec/2_related}
\input{sec/3_method}

\input{sec/4_exp}
\input{sec/5_con}
\input{sec/6_thank}
{
    \small
    \bibliographystyle{ieeenat_fullname}
    \bibliography{main}
}


\end{document}

%% file: sec/0_abstract.tex
\begin{abstract}
No-Reference Point Cloud Quality Assessment (NR-PCQA) still struggles with generalization, primarily due to the scarcity of annotated point cloud datasets. Since the Human Visual System (HVS) drives perceptual quality assessment independently of media types, prior knowledge on quality learned from images can be repurposed for point clouds. This insight motivates adopting Unsupervised Domain Adaptation (UDA) to transfer quality-relevant priors from labeled images to unlabeled point clouds. However, existing UDA-based PCQA methods often overlook key characteristics of perceptual quality, such as sensitivity to quality ranking and quality-aware feature alignment, thereby limiting their effectiveness.
To address these issues, we propose a novel Quality-aware Domain adaptation framework for PCQA, termed QD-PCQA. The framework comprises two main components: i) a Rank-weighted Conditional Alignment (RCA) strategy that aligns features under consistent quality levels and adaptively emphasizes misranked samples to reinforce perceptual quality ranking awareness; and ii) a Quality-guided Feature Augmentation (QFA) strategy, which includes quality-guided style mixup, multi-layer extension, and dual-domain augmentation modules to augment perceptual feature alignment.
Extensive cross-domain experiments demonstrate that QD-PCQA significantly improves generalization in NR-PCQA tasks. 
\end{abstract}

%% file: sec/1_intro.tex
\section{Introduction}
\label{sec:intro}
Point clouds, as 3D representations of objects or scenes \cite{chen2025mugsqa}, are widely used in applications such as Virtual Reality (VR) \cite{zhang2022openpointcloud}, Augmented Reality (AR) \cite{guo2020deep}, 3D modeling \cite{mekuria2016design}, and autonomous driving \cite{chen20203d}. However, point clouds are inevitably degraded during acquisition, processing, storage, and transmission \cite{gao2023opendmc}. Therefore, accurate Point Cloud Quality Assessment (PCQA) metrics are crucial.

\begin{figure}[t]
\centering
\includegraphics[width=\linewidth]{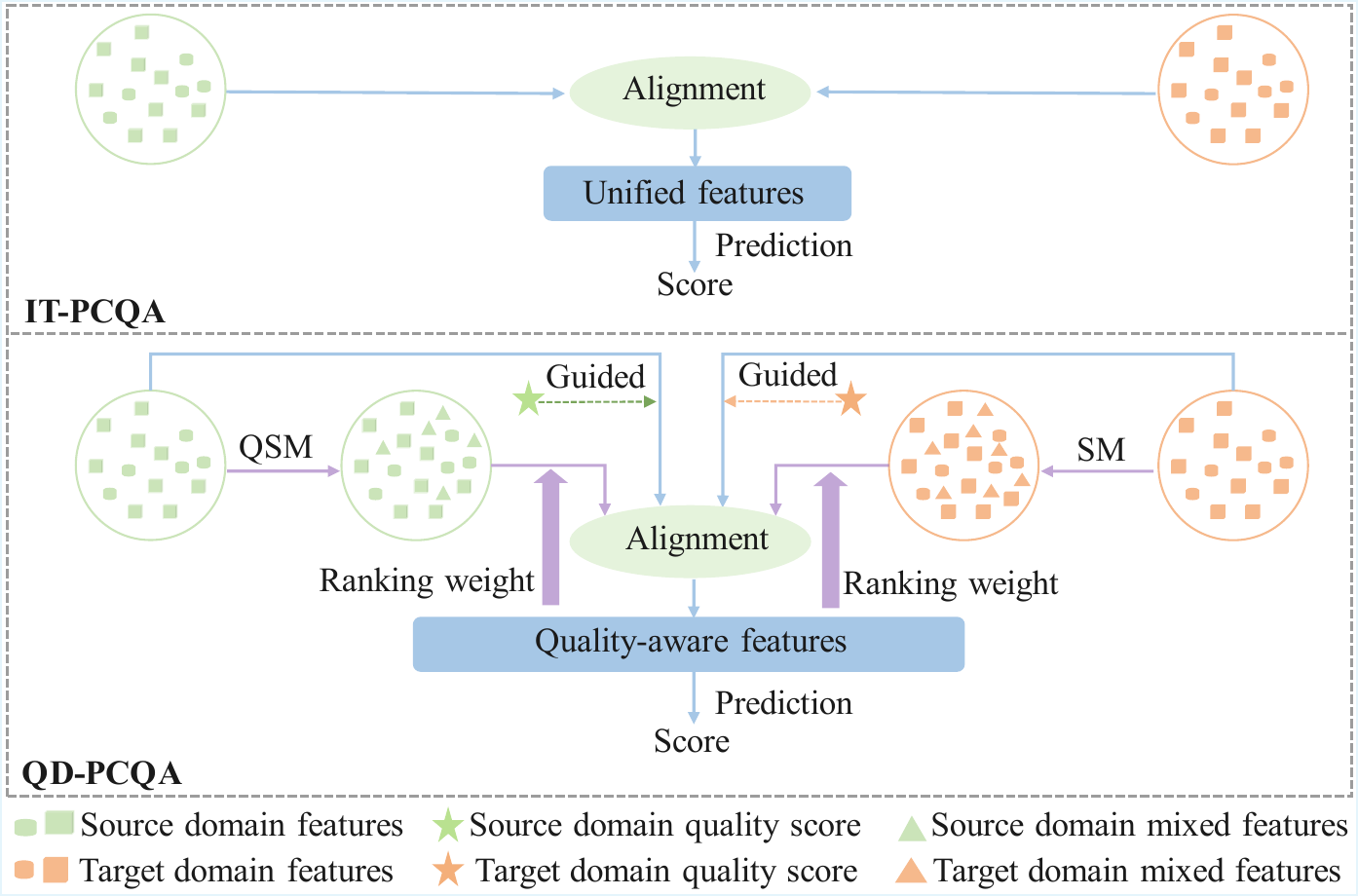}
\caption{
Comparison with IT-PCQA. IT-PCQA performs feature alignment via UDA, overlooking the characteristics of quality perception. In contrast, our QD-PCQA introduces two quality-aware strategies. The RCA strategy aligns features guided by quality scores of two domains, promoting quality-consistent adaptation. The QFA strategy enriches feature diversity through a multi-layer QSM and SM. Additionally, a rank-weighted module emphasizes misranked feature pairs to mitigate ranking bias.
}
\label{fig:1}
\end{figure}
PCQA metrics are commonly classified into Full-Reference (FR), Reduced-Reference (RR), and No-Reference (NR) methods. NR methods evaluate quality independently, without the original reference point cloud, whereas FR and RR methods rely on complete or partial reference information. Unfortunately, reference point clouds are often unavailable in practice due to transmission and storage constraints \cite{tangchina}. Therefore, NR methods are critical and are typically categorized into handcrafted and learning-based schemes \cite{liu2023jnmr,tang2024seeclear}. The former rely on manually designed features (\emph{e.g.,} geometry distortion, color deviation, and statistical descriptors), while the latter are data-driven models \cite{11419157,yang2026scaling}. Benefiting from deep learning, data-driven methods achieve better performance and are more promising \cite{liu2022temporal,jin2023kernel, jin2025feature}. However, compared with traditional vision tasks such as image classification, object detection, and even Image Quality Assessment (IQA), the annotated PCQA data are scarce, which restricts generalization abilities \cite{yang2022no}.


Unsupervised Domain Adaptation (UDA) \cite{xu2023universal} provides effective solutions to address the generalization challenge caused by limited labeled data by transferring knowledge from a labeled source domain to an unlabeled target domain, which has demonstrated its effectiveness on typical vision tasks, especially for image classification tasks. Generally, a classical UDA method directly aligns feature representations across source and target domains to enable knowledge transfer without target labels \cite{jiang2022transferability}.
Inspired by this, several works attempted to apply the UDA to the Quality Assessment (QA) tasks. For instance, Chen \emph{et al.} \cite{chen2020no} conducted the first attempt that utilized Maximum Mean Discrepancy (MMD) \cite{rozantsev2018beyond} to achieve UDA for the IQA task. Given that the Human Visual System (HVS) governs perceptual quality assessment independent of the specific media types (\emph{e.g.,} image, point cloud, or 3D mesh), Yang \emph{et al.} \cite{yang2022no} proposed IT-PCQA, which transfers the capability of quality prediction from images to point clouds, benefiting from the availability of large-scale annotated image datasets. Specifically, the features of images and point clouds were directly aligned to a unified distribution through adversarial-based Domain Adversarial Neural Network (DANN) \cite{ganin2016domain}, and image labels were used for point cloud quality regression training to alleviate the problem of insufficient point cloud quality labels. Meanwhile, Lu \emph{et al.} proposed the StyleAM \cite{lu2024styleam}, which introduced Style Mixup (SM) to augment source features and incorporates a style feature space into the alignment process, leading to further performance gains.

However, these methods inherit feature alignment strategies from image classification tasks, which focus on semantic consistency but ignore quality information. This may lead to unexpected \textbf{quality-regardless feature alignment}, where features with similar semantics but different quality levels are incorrectly aligned. For example, mapping high-quality ``tree'' images in the source domain to low-quality ``tree'' images in the target domain. Such misalignment compromises the model’s ability to distinguish perceptual quality, limiting its effectiveness in quality assessment tasks.
Besides, the strength of feature alignment is always unchanged, which should also be refined. Moreover, although StyleAM \cite{lu2024styleam} improves performance through feature augmentation, it introduces several additional issues. First, \textbf{quality-regardless feature augmentation}: SM performs random interpolation of style features without considering quality information, resulting in augmented features that fail to represent perceptual quality effectively.
Second, \textbf{layer-regardless feature augmentation}: Applying augmentation solely to the final layer overlooks the complementary nature of hierarchical features. Features from shallow layers are more sensitive to low-level distortions relevant to high-quality images, while deeper layers capture high-level semantics critical for assessing lower-quality samples. Ignoring this variation reduces the representational richness of the augmented features for quality assessment. Third, \textbf{augmentation imbalance}: As only features of the source domain are augmented, which widens the domain gap, making it easy to make the discriminator to distinguish the two domains, thereby weakening the adversarial feature learning of quality representations. 

To overcome the above problems, we propose a novel QD-PCQA as shown in Figure~\ref{fig:1}, consisting of two main components: i) a \textbf{Rank-weighted Conditional Alignment (RCA)} strategy, which includes a quality-aware conditional module and a rank-weighted module. 
The former module introduces a quality-biased feature alignment method, which uses ground-truth quality scores from the source domain and pseudo quality scores from the target domain as condition information. By aligning features with similar quality levels, it addresses the quality-regardless feature alignment issue. 
The latter module assigns greater weights to misranked feature pairs to strengthen their alignment constraints, leading to more accurate correction of ranking bias. ii) a \textbf{Quality-guided Feature Augmentation (QFA)} strategy, which includes a Quality-guided Style Mixup (QSM) module, a multi-layer extension module, and a dual-domain augmentation module. 
The QSM module integrates a novel quality-guided selection and an SM operation. 
The selection operation matches samples with similar quality levels, ensuring that the subsequent SM produces quality-consistent and quality-aware augmentations to solve the quality-regardless feature augmentation issue.
The multi-layer extension module is designed to address the issue of layer-regardless feature augmentation by leveraging the complementary nature of hierarchical features. To this end, it first categorizes input samples into high, medium, and low quality groups by a quality stratifier, and then embeds the QSM module into the shallow, middle, and deep layers of the network, respectively. The dual-domain augmentation module applies the QSM-based multi-layer extension module in the source domain and SM in the target domain integrated into the adversarial-based DANN \cite{ganin2016domain}, thereby solving the augmentation imbalance problem. 

The contributions can be summarized as follows:
\begin{itemize}
\item We propose a novel domain adaptation quality assessment framework called QD-PCQA that leverages prior quality knowledge from images to predict point cloud quality. 
\item We develop the RCA strategy that aligns features under consistent quality levels to solve the issue of quality-regardless feature alignment, and adaptively emphasizes misranked samples to refine alignment strength, thereby enhancing quality ranking sensitivity.
\item  We develop the QFA strategy that incorporates QSM, multi-layer feature integration, and dual-domain augmentation, enabling hierarchical quality representation and augmenting perceptual feature alignment.
\end{itemize}

%% file: sec/2_related.tex
\section{Related Work}
\label{sec:related}
\subsection{Point Cloud Quality Assessment}
Early Point Cloud Quality Assessment (PCQA) research primarily focused on FR metrics to support Point Cloud Compression (PCC). MPEG standardized several FR metrics, including point-to-point \cite{mekuria2016evaluation}, point-to-plane \cite{tian2017geometric}, and PSNRyuv \cite{torlig2018novel}. To better align with human perception, later methods incorporated Human Visual System (HVS)-based features, such as structural similarity \cite{yang2020inferring}, curvature \cite{he2021towards}, and color brightness \cite{meynet2020pcqm}.
However, in the absence of reference point clouds, NR-PCQA has attracted increasing attention. Chetouani \cite{chetouani2021deep} proposed a Convolutional Neural Network (CNN)-based method using local geometric features. Liu \emph{et al.} \cite{liu2021pqa} enhanced representation through multi-view projections. Although 2D projections may introduce some distortions and the number and angles of views can affect quality representation, they are commonly used for subjective rating, and most PCQA metrics adopt projected images or videos as inputs \cite{yang2022no}. Yang \emph{et al.} further bridged 2D and 3D features by adaptively weighting projection blocks according to their relative contributions, showing through subjective experiments that point cloud quality can be effectively captured using six orthogonal views\cite{yang2022no}.

The above methods require large-scale training data from the same distribution as testing, limiting generalization to unseen scenarios. Hence, we propose a UDA-based method that transfers quality prediction capability from images to point clouds, considering that HVS drives perceptual assessment independently of media types.
\subsection{Unsupervised Domain Adaptation}
UDA aims to mitigate domain shift between labeled source and unlabeled target domains \cite{rozantsev2018beyond,ganin2016domain}. Existing methods are typically categorized into marginal, conditional, and joint distribution alignment. Marginal alignment only focuses on global feature consistency using techniques such as MMD \cite{rozantsev2018beyond} or adversarial learning (\emph{e.g.,} Domain Adversarial Neural Network (DANN) \cite{ganin2016domain}). Joint alignment \cite{long2017deep,long2018conditional} models feature–label dependencies, while conditional alignment further leverages label information, \emph{e.g.,} using the Conditional Operator Discrepancy (COD) \cite{yang2024cod}. 
Recently, UDA has also been explored in quality assessment tasks.
Chen \emph{et al.} \cite{chen2020no} first applied MMD \cite{rozantsev2018beyond} for feature alignment in IQA. StyleAM \cite{lu2024styleam} introduced a style space into DANN-based adversarial alignment \cite{ganin2016domain}. IT-PCQA \cite{yang2022no} employed DANN \cite{ganin2016domain} to unify image and point cloud features.

However, existing UDA-based PCQA methods often overlook the characteristics of quality perception, leading to sub-optimal results for NR-PCQA tasks. To address this issue, we propose the novel QD-PCQA.

%% file: sec/3_method.tex
\section{Method}
\label{sec:method}
\subsection{Data Preprocessing Module}
We project 3D point cloud data onto the six vertical sides of a cube, generating multiple side views, following IT-PCQA \cite{yang2022no} to share a common feature extractor. These views are then stitched into a multi-view image and, along with natural images, resized to \(224 \times 224\) for unified input to the subsequent feature extractor.
\subsection{QFA Strategy}
The proposed Quality-guided Feature Augmentation (QFA) strategy enhances feature generalization by extracting domain-invariant and quality-aware representations to support both prediction and alignment. Built upon a modified ResNet-50 \cite{he2016deep} backbone (excluding the original classification head), the QFA strategy integrates three key components, including Quality-guided Style Mixup (QSM), multi-layer extension, and dual-domain augmentation modules. 
To extract deep representations for subsequent prediction and alignment, we first obtain the feature $f_s^i \in \mathbb{R}^{C \times H \times W}$ from the source domain and $f_t^j \in \mathbb{R}^{C \times H \times W}$ from the target domain, formulated as:
\begin{equation}
f_s^i = G(x_s^i), \quad f_t^j = G(x_t^j),
\end{equation}
where \(x_s^i\) denotes the $i$-th source domain sample and \(x_t^j\) denotes the $j$-th target domain sample; $G(\cdot)$ denotes the feature extractor; \emph{C}, \emph{H} and \emph{W} denote the channel, height, and width of features, respectively. 
\subsubsection{QSM Module}
\begin{figure*}[t]
\centering
\includegraphics[width=\textwidth]{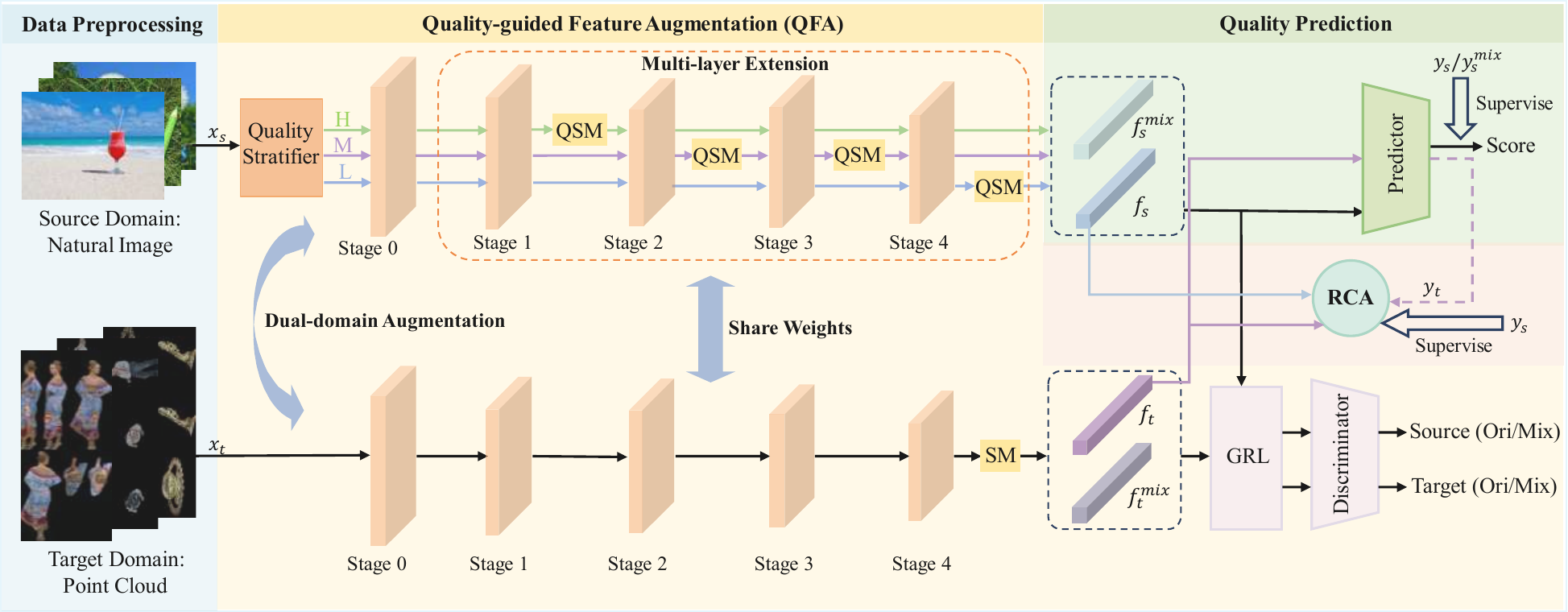}
\caption{ Architecture of the proposed QD-PCQA. Given the source-domain image \(x_s\) and target-domain point cloud \(x_t\), both are cropped to the same size. Then processed by QFA for quality-guided feature augmentation and by RCA for rank-aware conditional alignment. Finally, the predictor outputs the final quality score.
}
\label{fig:2}
\end{figure*}
To solve the issue of quality-regardless feature augmentation, the Quality-guided Style Mixup (QSM) module is proposed, which integrates a novel quality-guided selection and a Style Mixup (SM) operation. Unlike conventional SM that randomly mixes samples, the proposed selection operation leverages a Gaussian kernel \cite{yao2022c} to adaptively pair source samples with similar quality scores, thereby guiding the SM operation to preserve quality consistency and enhance quality-aware feature augmentation.

Specifically, to ensure that source domain samples with similar perceptual quality are more likely to be selected for mixing, we define the selection probability of a candidate pair $(x_s^{i^*}, y_s^{i^*})$ given a sample $(x_s^i, y_s^i)$, formulated as:
\begin{equation}
P((x_s^{i^*}, y_s^{i^*}) | (x_s^i, y_s^i)) \propto \exp\left( -\frac{(y_s^i - y_s^{i^*})^2}{2\tau^2} \right),
\end{equation}
where $\tau$ represents the bandwidth parameter of the Gaussian kernel, and $y_s^i$ and $y_s^{i^*}$ represent the labels of source domain samples $i$ and ${i^*}$, respectively.

After pairing, SM is applied to generate mixed features. Specifically, we compute the channel-wise mean $u(f)_c$ and standard deviation $\sigma(f)_c$ of a feature $f \in \mathbb{R}^{C \times H \times W}$, formulated as:
\begin{gather}
u(f)_c = \frac{1}{HW} \sum_{h=1}^{H} \sum_{w=1}^{W} f_{c,h,w}
,\notag \\
\sigma(f)_c = \sqrt{\frac{1}{HW} \sum_{h=1}^{H} \sum_{w=1}^{W} \left( f_{c,h,w} - u(f)_c \right)^2,}
\end{gather}
where $f_{c,h,w}$ denotes the feature at channel $c$, height $h$ and width $w$; $H$ and $W$ denote the height and width of the feature, respectively.

Given the feature styles $<u(f_s^i), \sigma(f_s^i)>$ and $< u(f_s^{i^*}) , \sigma(f_s^{i^*}) >$ of two samples, along with their corresponding quality scores $y_s^i$ and $y_s^{i^*}$, we construct a mixing vector $\lambda$ to blend the feature styles and quality labels. The mixed feature styles $u(f)^{\text{mix}}$ and $\sigma(f)^{\text{mix}}$, as well as the mixed label $y_s^{\text{mix}}$, formulated as respectively:
\begin{align}
u(f)^\text{mix} &= \lambda u(f_s^i) + (1 - \lambda) u(f_s^{i^*}),
\notag \\
\sigma(f)^\text{mix} &= \lambda \sigma(f_s^i) + (1 - \lambda) \sigma(f_s^{i^*}), \notag \\
y_s^\text{mix} &= \lambda y_s^i + (1 - \lambda) y_s^{i^*},
\end{align}
where $\lambda \sim \text{Beta}(\alpha, \alpha)$ denotes sampled from the Beta distribution with hyperparameter $\alpha > 0$.

Finally, the mixed style feature $f_s^{\text{mix}}$ is formulated as:
\begin{equation} \label{eq:mix_final}
f_s^{\text{mix}} = \sigma(f)^{\text{mix}} \frac{f_s^i - u(f_s^i)}{\sigma(f_s^i)} + u(f)^{\text{mix}}.
\end{equation}

\subsubsection{Multi-Layer Extension Module}
To solve the issue of layer-regardless feature augmentation, the multi-layer extension module is proposed to determine at which layer to use the QSM based on the quality scores of the source domain samples. As illustrated in Figure~\ref{fig:2}, we introduce a quality stratifier that categorizes input source domain samples into high, medium, and low quality groups. It follows a quantile-based quality classification method \cite{liu2017rankiqa}. Based on this stratification, the multi-layer extension module embeds the QSM module into the shallow, middle, and deep layers of the network, respectively, considering the complementary nature of hierarchical features to solve the layer-regardless feature augmentation issue. Specifically, QSM is selectively applied at different network stages based on image quality. Stage 1 is used for high-quality samples, stages 2–3 for medium-quality samples, and stage 4 for low-quality samples.


Existing methods suffer from layer-indiscriminate feature augmentation because they overlook the hierarchical complementarity of deep features in quality assessment \cite{chen2024topiq}. Shallow layers are more sensitive to low-level distortions (\emph{e.g.,} slight blur or blocking artifacts) typical of high-quality samples, while deeper layers capture high-level semantics crucial for assessing severely degraded ones \cite{zhou2026high}. Consequently, the reliance on different feature layers varies with distortion severity and perceived quality: high-quality samples depend more on shallow features, whereas low-quality samples rely on deeper representations.
To overcome this, we design a quality-aware hierarchical augmentation mechanism that applies QSM according to the feature-level relevance to perceptual quality, rather than uniformly across layers.


\subsubsection{Dual-Domain Augmentation Module}
To address the augmentation imbalance, a dual-domain augmentation module is designed to enable feature augmentation in both domains, rather than only in the source domain. The multi-layer extension module is applied to the source domain, while SM is applied to target features after stage 4, as target quality scores are unavailable.
The module enriches feature representations and promotes greater overlap between the source and target feature distributions, facilitating the learning of shared, domain-invariant representations and thus narrowing the domain gap. Moreover, it increases the difficulty of the domain discriminator, encouraging the feature extractor to learn more robust and invariant features. Since the target domain lacks labels, we perform uniform feature mixing at the final layer without mixing labels, aiming solely to enhance representation. The mixed features from both domains are finally fed into a DANN \cite{ganin2016domain} through a Gradient Reversal Layer (GRL) for adversarial training.

We can optimize the discriminator \emph{D} with a binary cross entropy loss $\mathcal{L}_D$, which is formulated as:
\begin{equation}
\mathcal{L}_D = -\frac{1}{n_s} \sum_{i=1}^{n_s} \log(1 - D(f_s^i)) - \frac{1}{n_t} \sum_{j=1}^{n_t} \log(D(f_t^j)),
\end{equation}
where $n_s$ and $n_t$ denote the number of source domain and target domain samples, $f_s^i$ and $f_t^j$ are the feature representations from the source and target domains, and $D(\cdot)$ is the domain discriminator.
\subsection{RCA Strategy}
To solve the issue of quality-regardless feature alignment, we design a Rank-weighted Conditional Alignment (RCA) strategy, which includes a quality-aware conditional module and a rank-weighted module. The quality-aware conditional module is a feature alignment with quality-aware bias, which uses the true quality scores from the source domain and pseudo quality scores from the target domain as conditions to align features with a consistent quality level. The rank-weighted module assigns greater weights to misranked feature pairs to strengthen their alignment constraints, leading to more accurate correction of ranking bias. 

Specifically, the RCA strategy is built on the statistical metric of COD \cite{yang2024cod}, which measures local distribution differences in a fine-grained manner. However, COD \cite{yang2024cod} assigns equal weights to all sample pairs, ignoring the role of quality ranking in perceptual alignment. To address this issue, RCA introduces a rank-weighted module to highlight sample pairs that exhibit ranking bias in cross-domain prediction. Compared to other sample pairs, these positions are more likely to reflect domain differences, thus increasing their weight prompts the model to prioritize correcting quality-perceptual alignment errors. 
The following loss function $\mathcal{L}_{R}$ is designed to further optimize the feature extractor \emph{G} formulated as:
\begin{equation}
\begin{split}
\mathcal{L}_{R} &= \operatorname{tr}\!\left( \mathbf{K}_Y^{tt} (\mathbf{K}_Y^{tt}+ \epsilon \mathbf{I})^{-1} \tilde{\mathbf{K}}_X^{tt} (\mathbf{K}_Y^{tt}+ \epsilon \mathbf{I})^{-1} \right) \\
& \quad +\operatorname{tr}\!\left( \mathbf{K}_Y^{ss} (\mathbf{K}_Y^{ss}+ \epsilon \mathbf{I})^{-1} \tilde{\mathbf{K}}_X^{ss} (\mathbf{K}_Y^{ss}+ \epsilon \mathbf{I})^{-1} \right) \\
& \quad - 2\operatorname{tr}\!\left( \mathbf{K}_Y^{ts} (\mathbf{K}_Y^{ss}+ \epsilon \mathbf{I})^{-1} \tilde{\mathbf{K}}_X^{st} (\mathbf{K}_Y^{tt}+ \epsilon \mathbf{I})^{-1} \right),
\end{split}
\end{equation}
where $\operatorname{tr}(\cdot)$ denotes the trace. 
$\mathbf{K}_Y^{tt}$ and $\mathbf{K}_Y^{ss}$ are the label kernel matrices for the target and source domains, and $\mathbf{K}_Y^{ts}$ is the cross-domain label kernel matrix. 
$(\mathbf{K} + \epsilon \mathbf{I})^{-1}$ denotes the Tikhonov-regularized inverse \cite{yang2024cod} of $\mathbf{K}$ for numerical stability, where $\mathbf{I}$ is the identity matrix and $\epsilon > 0$ is a regularization constant. $\tilde{\mathbf{K}}_X^{tt}$, $\tilde{\mathbf{K}}_X^{ss}$, and $\tilde{\mathbf{K}}_X^{st}$ are the feature kernel matrices constructed with our proposed rank-weighted strategy. For example, $\tilde{\mathbf{K}}_X^{st}(i,j)$ denotes the rank-weighted feature kernel value between the $i$-th source sample and the $j$-th target sample, which is computed as follows:
\begin{equation}
\tilde{\mathbf{K}}_X^{st}(i, j) = k({f}_s^i, {f}_t^j) \cdot (1 + \mathbf{W}^{st}(i, j)),
\end{equation}
where ${k({f}_s^i,{f}_t^j)}$ denotes the kernel function (\emph{i.e.,} Gaussian kernel) to measure the similarity between source-domain feature ${f}_s^i$ and target-domain feature ${f}_t^j$. $\mathbf{W}^{st}(i, j)$  denotes the rank-weight matrix element, to capture the contribution of the sample pair $(i,j)$, formulated as:
\begin{equation}
\mathbf{W}^{st}(i, j) = \max\big(0, -(\hat{y}_s^i - \hat{y}_t^j) \cdot \operatorname{sign}(y_s^i - y_t^j)\big),
\end{equation}
where $\hat{y}_s^i$ denotes the predicted value for the $i$-th sample in the source domain;
$\hat{y}_t^j$ denotes the predicted value for the $j$-th sample in the target domain; $y_s^i$ represents the true label of the $i$-th sample in the source domain;
$y_t^j$ represents the pseudo label of the $j$-th sample in the target domain;
$\operatorname{sign}(\cdot)$ denotes the sign function, which returns $1$ for positive inputs, $-1$ for negative inputs, and $0$ for zero.

Since early pseudo-labels are unreliable due to the unconstrained model, directly applying RCA may lead to incorrect alignment. Therefore, we propose a two-stage training strategy to introduce the reliable pseudo-label after the model is constrained and improve the robustness of RCA, as detailed in Section~\ref{sec:training_strategy}.
\subsection{Quality Prediction Module}
We employ a quality predictor \emph{P} with two Fully-Connected layers to regress quality scores from the aligned features.

Given a set of distorted source images with features and quality labels, denoted as $\{x_s^i | f_s^i, y_s^i\}_{i=1}^{n_s}$, where $f_s^i = G(x_s^i)$, the loss function $\mathcal{L}_{P}$ of the prediction network is formulated as:
\begin{equation}
\mathcal{L}_{P} = \frac{1}{n_s} \sum_{i}^{n_s} \left( \hat{y}_s^i - y_s^i \right)^2,
\end{equation} 
where $y_s^i$ denotes the quality labels of the distorted samples and  $\hat{y_{s}^i} = P(f_s^i)$ denotes the predicted quality score of samples in the source domain. 

Note that during training, to enable RCA, pseudo-labels for target domain samples are required. Therefore, target features $f_t^j = G(x_t^j)$ are also passed through the predictor \emph{P} for quality estimation. At test time, the quality score of a target sample $x_t^j$ is predicted in the same way as during pseudo-label $y_t$ generation.
\subsection{Training Strategy}
\label{sec:training_strategy}
After determining the network structure, we adopt a two-stage training strategy to progressively learn domain-invariant features and stabilize quality prediction. Due to the large domain gap, pseudo-labels are unreliable in the early stage. Thus, we first train a DANN-based model without pseudo-labels to achieve initial feature alignment. Once the model stabilizes, we introduce the pseudo-label–dependent RCA strategy to further refine cross-domain alignment and quality regression. As shown in Table~\ref{tab:6} and Figure~\ref{fig:domain}, this strategy yields more stable convergence and better performance. 
Notably, the RCA strategy is applied solely to the original source and target features to ensure accurate distribution alignment.

To enable stochastic application of the QSM and SM modules, we define two loss formulations:

The mixed-sample loss, denoted as $\mathcal{L}^{\text{mix}}_{all}$, is computed when QSM and SM are applied, formulated as:
\begin{equation} 
\begin{aligned} 
\mathcal{L}_\text{all}^\text{mix} &= \lambda_1 \mathcal{L}_P(\hat{y}_s^\text{mix}, y_s^\text{mix}) 
+ \lambda_2 \mathcal{L}_D(f_s^\text{mix}, f_t^\text{mix}) \\
&+ \lambda_3 \mathcal{L}_R(y_s, y_t,f_s,f_t),
\end{aligned}
\end{equation}

The original-sample loss, denoted as $\mathcal{L}^{\text{orig}}_{\text{all}}$, is used when no mixing is applied, formulated as:
\begin{equation}
\begin{aligned}
\mathcal{L}_\text{all}^\text{orig} &= \lambda_1 \mathcal{L}_P(\hat{y}_s, y_s) + \lambda_2 \mathcal{L}_D(f_s, f_t) \\
&\quad + \lambda_3 \mathcal{L}_R(y_s, y_t,f_s,f_t)
\end{aligned}
\end{equation}
where $\lambda_1$, $\lambda_2$, and $\lambda_3$ denote loss weights; where we use $\mathcal{L}{\text{all}}^{\text{mix}}$ if $p > 0.5$, and $\mathcal{L}{\text{all}}^{\text{orig}}$ otherwise.
$p$ represents a draw from a uniform distribution over $[0,1]$, indicating that QSM and SM are applied with a probability of $0.5$.

%% file: sec/4_exp.tex
\section{Experiments}
\label{sec:exp}
\begin{table*}[t]
\caption{Performance of training on TID2013 \cite{ponomarenko2015image} and testing on SJTU-PCQA \cite{yang2020predicting} or WPC \cite{su2019perceptual}.}
\label{tab:1}
  \centering
  \small
  \begin{tabular}{l|c|cccc|cccc}
    \toprule
    \multirow{2}{*}{\centering Methods} & \multirow{2}{*}{\centering Model} 
    & \multicolumn{4}{c|}{TID2013 \cite{ponomarenko2015image} \(\to\)SJTU-PCQA \cite{yang2020predicting}} & \multicolumn{4}{c}{TID2013 \cite{ponomarenko2015image} \(\to\)WPC \cite{su2019perceptual} } \\
    \cmidrule(lr){3-6} \cmidrule(lr){7-10}
    & & PLCC↑ & SROCC↑ & KROCC↑ & RMSE↓ & PLCC↑ & SROCC↑ & KROCC↑ & RMSE↓ \\
    \midrule
    StyleAM \cite{lu2024styleam} & I-to-I & 0.612 & 0.519 & 0.352 & 1.883 & 0.352 & 0.299 & 0.265 & 21.323 \\
    Chen’s \cite{chen2020no}& I-to-I & 0.706 & 0.632 & 0.453 & 1.667 & 0.386 & 0.302 & 0.293 & 21.102 \\
    DANN \cite{ganin2016domain}& I-to-PC & 0.596 & 0.512 & 0.489 & 1.965 & 0.325 & 0.296 & 0.245 & 21.364 \\
    COD \cite{yang2024cod} & I-to-PC & 0.712 & 0.611 & 0.492 & 1.735 & 0.426 & 0.396 & 0.320 & 20.994 \\
    No Adapt  & I-to-PC & 0.548 & 0.444 & 0.321 & 1.995 & 0.320 & 0.296 & 0.221 & 21.423 \\
    IT-PCQA \cite{yang2022no}& I-to-PC & 0.693 & 0.636 & 0.493 & 1.624 & 0.429 & 0.403 & 0.323 & 20.867 \\
    QD-PCQA (Ours) & I-to-PC & \textbf{0.842} & \textbf{0.753} & \textbf{0.566} & \textbf{1.358} & \textbf{0.563} & \textbf{0.572} & \textbf{0.462} & \textbf{20.835} \\
    \bottomrule
  \end{tabular}
\end{table*}
\renewcommand{\arraystretch}{1.2}
\begin{table*}[htbp]
\caption{Performance of training on KADID-10k \cite{lin2019kadid} and testing on SJTU-PCQA \cite{yang2020predicting} or WPC \cite{su2019perceptual}.}
\label{tab:2}
  \centering
  \small
  \begin{tabular}{l|c|cccc|cccc}
    \toprule
    \multirow{2.6}{*}{\centering Methods} & \multirow{2.6}{*}{\centering Model} 
    & \multicolumn{4}{c|}{KADID-10k \cite{lin2019kadid}\(\to\)SJTU-PCQA \cite{yang2020predicting}} & \multicolumn{4}{c}{KADID-10k \cite{lin2019kadid}\(\to\)WPC \cite{su2019perceptual} } \\
    \cmidrule(lr){3-6} \cmidrule(lr){7-10}
    & & PLCC↑ & SROCC↑ & KROCC↑ & RMSE↓ & PLCC↑ & SROCC↑ & KROCC↑ & RMSE↓ \\
    \midrule
    StyleAM \cite{lu2024styleam} & I-to-I & 0.698 & 0.589 & 0.450 & 1.712 & 0.378 & 0.320 & 0.267 & 21.303 \\
    Chen’s \cite{chen2020no}& I-to-I & 0.643 & 0.568 & 0.406 & 1.824 & 0.402 & 0.312 & 0.303 & 20.991 \\
    DANN \cite{ganin2016domain} & I-to-PC & 0.602 & 0.534 & 0.446 & 1.812 & 0.324 & 0.284 & 0.256 & 21.385 \\
    COD \cite{yang2024cod} & I-to-PC & 0.721 & 0.612 & 0.496 & 1.703 & 0.423 & 0.395 & 0.315 & 20.965 \\
    No Adapt & I-to-PC & 0.552 & 0.462 & 0.392 & 1.965 & 0.321 & 0.297 & 0.254 & 21.403 \\
    IT-PCQA \cite{yang2022no} & I-to-PC & 0.703 & 0.641 & 0.512 & 1.606 & 0.432 & 0.402 & 0.386 & 20.945 \\
    QD-PCQA (Ours) & I-to-PC & \textbf{0.843} & \textbf{0.724} & \textbf{0.528} & \textbf{1.305} & \textbf{0.553} & \textbf{0.534} & \textbf{0.468} & \textbf{20.794} \\
    \bottomrule
  \end{tabular}
  
\end{table*}

\subsection{Experimental Setup}
\subsubsection{Dataset and Evaluation Metrics}
We evaluate our method on four datasets, including TID2013 \cite{ponomarenko2015image}  and KADID-10k \cite{lin2019kadid} for natural images, and SJTU-PCQA \cite{yang2020predicting} and WPC \cite{su2019perceptual} for point clouds. We adopt Pearson’s Linear Correlation Coefficient (PLCC), Spearman Rank Order Correlation Coefficient (SROCC), Kendall Rank Order Correlation Coefficient (KROCC), and Root Mean Square Error (RMSE) as metrics. Better performance is indicated by higher PLCC, SROCC, and KROCC, and lower RMSE.
\subsubsection{Implementation Details}
All experiments are implemented in PyTorch on an NVIDIA RTX 4090 GPU. Natural image datasets serve as source domains, while point cloud datasets are split into 75\% for training and 25\% for testing as target domains. We adopt a modified ResNet-50 \cite{he2016deep} backbone pretrained on ImageNet \cite{deng2009imagenet}. The domain discriminator $D$ consists of two Fully-Connected(FC) layers with ReLU activations and one FC layer with Sigmoid. The predictor uses two FC-ReLU layers followed by a Sigmoid function. Input images are randomly cropped and flipped during training, and center-cropped during testing.
We use the SGD optimizer with momentum 0.9, weight decay $5 \times 10^{-4}$, a batch size of 36, and 30,000 iterations. DANN \cite{ganin2016domain} is applied during the first 5,000 iterations (warm-up), followed by joint optimization with RCA for the remaining iterations.
For QSM, $\alpha$ is determined via ablation studies, set to 1 for TID2013 \cite{ponomarenko2015image} and 0.5 for KADID-10k \cite{lin2019kadid} to augment style diversity. 
Finally, we empirically set $\lambda_1 = \lambda_2 = \lambda_3$ to balance the three loss terms, the Gaussian kernel bandwidth to $\tau = 5 \!\times\! 10^{-2}$ for stable optimization, and the Tikhonov regularization \cite{yang2024cod} parameter to $\epsilon = 10^{-3}$. Samples are divided into three quality levels (low, medium, high) based on the 33\% and 67\% quantiles of subjective scores.
\renewcommand{\arraystretch}{1.2} 

\subsection{Performance Evaluation}
\begin{figure}[t]
\centering
\includegraphics[width=0.9\linewidth]{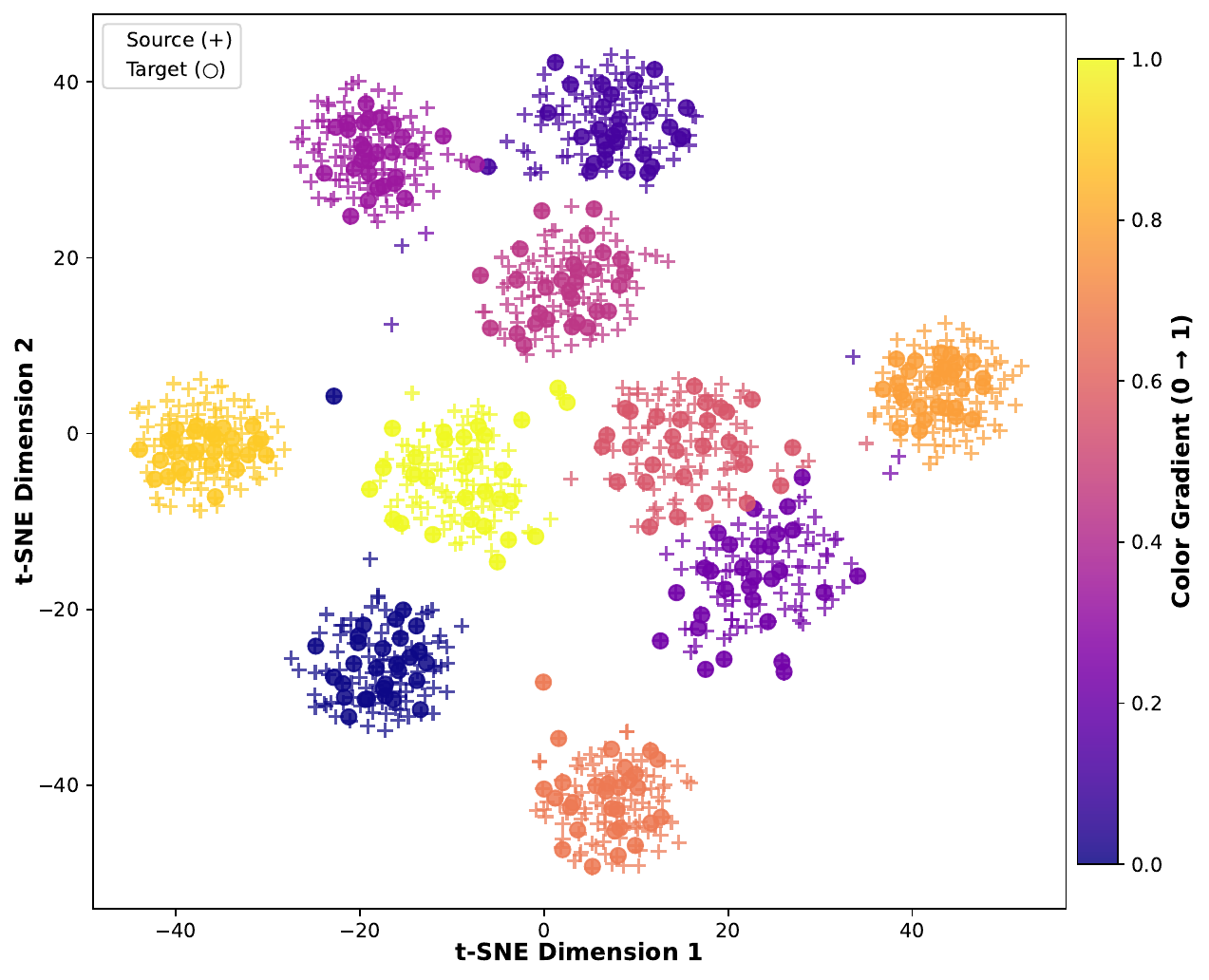}
\caption{t-SNE visualization of the aligned representations learned from the source and target domains. Label values are denoted by color gradients. Ten label values are selected from the range of the variable for visualization.}
\label{fig:domain}
\end{figure}
Because images and point cloud projections share the same 2D format, image-to-image IQA transfer methods can also be applied to image-to-point cloud transfer.
Therefore, we compare the proposed QD-PCQA with three categories of reproduced baselines: image-to-image transfer IQA methods (\emph{e.g.}, StyleAM \cite{lu2024styleam}, Chen’s \cite{chen2020no}), general UDA methods adapted for image-to-point cloud transfer (\emph{e.g.}, DANN \cite{ganin2016domain}, COD \cite{yang2024cod}), and image-to-point cloud transfer methods such as IT-PCQA \cite{yang2022no}. 
``No Adapt'' denotes our reproduced baseline using ResNet-50 \cite{he2016deep} as the feature extractor, followed by a quality predictor.

As shown in Table~\ref{tab:1}, we compare QD-PCQA with representative baselines trained on TID2013 \cite{ponomarenko2015image} and evaluated on SJTU-PCQA \cite{yang2020predicting} and WPC \cite{su2019perceptual}. QD-PCQA consistently achieves state-of-the-art performance. For example, in the TID2013 \cite{ponomarenko2015image} → SJTU-PCQA \cite{yang2020predicting} setting, it achieves a PLCC of 0.842 and an RMSE of 1.358, surpassing IT-PCQA \cite{yang2022no} by 21.5\% in PLCC and reducing RMSE by 16.4\%. On TID2013 \cite{ponomarenko2015image} → WPC \cite{su2019perceptual}, it reaches a PLCC of 0.563, outperforming DANN \cite{ganin2016domain} by 73.2\% and IT-PCQA \cite{yang2022no} by 31.2\%, demonstrating strong robustness under challenging cross-domain conditions.
Similarly, Table~\ref{tab:2} shows results for baselines trained on KADID-10k \cite{lin2019kadid}. QD-PCQA consistently leads across evaluations. For KADID-10k \cite{lin2019kadid} → SJTU-PCQA \cite{yang2020predicting}, it achieves a PLCC of 0.843 and RMSE of 1.305, surpassing IT-PCQA \cite{yang2022no} by 19.9\% in PLCC and reducing RMSE by 18.7\%.

These results align with the intrinsic differences in dataset distortions. SJTU-PCQA \cite{yang2020predicting} primarily contains low-level texture and color degradations, often as single-type or binary distortions (\emph{e.g.,} color noise, downsampling), which are similar to pixel-level degradations in image datasets such as TID2013 \cite{ponomarenko2015image}. This similarity facilitates cross-domain transfer, explaining the generally higher performance on SJTU-PCQA \cite{yang2020predicting}. In contrast, WPC \cite{su2019perceptual} exhibits high-level semantic distortions caused by G-PCC compression, which differ significantly from image-domain degradations and pose greater challenges for cross-domain alignment. To address these challenges, our QFA strategy effectively enriches the source domain with diverse samples, facilitating shared feature learning and narrowing the domain gap. Meanwhile, the multi-layer extension module enhances texture details for high-quality samples and strengthens semantic robustness for low-quality ones. As shown in Figure~\ref{fig:domain}, QD-PCQA aligns source and target domain features effectively, confirming its generalization.

\subsection{Ablation Study}
This section assesses the effectiveness of each component using TID2013 \cite{ponomarenko2015image} as the source and SJTU-PCQA \cite{yang2020predicting} as the target domain.
\subsubsection{QSM Module}
We compare three style mixing strategies: No Style Mixup, Style Mixup, and our proposed QSM. As shown in Table~\ref{tab:3}, introducing style mixing improves cross-domain generalization, while QSM further boosts performance by guiding the mixing according to quality similarity, ensuring perceptual consistency in the augmented features. Moreover, the ablation on the Beta parameter $\alpha$ in Table~\ref{tab:4} shows that $\alpha = 1$ achieves the best results, highlighting that the mixing strength directly affects performance by balancing style diversification and quality structure preservation.

\begin{table}[H]  
 \caption{Ablation study on our QSM module.}
 \label{tab:3}
  \centering
  \small
  \begin{tabular}{lcccc}
    \toprule
    Methods & PLCC↑ & SROCC↑ & KROCC↑ & RMSE↓ \\
    \midrule
    No Style Mixup & 0.792 & 0.676 & 0.484 & 1.543\\
    Style Mixup & 0.820 & 0.711 & 0.507 & 1.392\\
    
    QSM & \textbf{0.842} & \textbf{0.753} & \textbf{0.566} & \textbf{1.358} \\
    \bottomrule
  \end{tabular}
 
\end{table}

\begin{table}[H]
  \centering
  
  \caption{Ablation study on the Beta distribution parameter $\alpha$, which controls the concentration of the mixing coefficient $\lambda$.}
  \label{tab:4}
  \small
  \begin{tabular}{lcccc} 
    \toprule
    $\alpha$ & PLCC$\uparrow$ & SROCC$\uparrow$ & KROCC$\uparrow$ & RMSE$\downarrow$ \\
    \midrule
    0.1            & 0.834          & 0.732           & 0.512           & 1.374            \\ 
    0.2            & 0.838          & 0.749           & 0.532           & 1.364            \\ 
    0.5            &0.840 & 0.752 & 0.556 & 1.361\\ 
    1.0 & \textbf{0.842} & \textbf{0.753} & \textbf{0.566} & \textbf{1.358}\\
    \bottomrule
  \end{tabular}
\end{table}

\subsubsection{Multi-Layer Extension Module}
We compare Style Mixup (SM) applied at different network stages and our proposed multi-layer extension. As shown in Table~\ref{tab:5}, the multi-layer extension consistently performs best, indicating that leveraging hierarchical features in a complementary way is more effective for PCQA.

\begin{table}[H]
\caption{Ablation study on SM insertion in different layers.}
\label{tab:5}
  \centering
 \small
  \begin{tabular}{lcccc}
    \toprule
    Methods & PLCC↑ & SROCC↑ & KROCC↑ & RMSE↓ \\
    \midrule
    @stage1 & 0.818 & 0.698& 0.493 & 1.412\\
    @stage4 & 0.815 & 0.701 & 0.496 & 1.381\\
    @stage2,3 & 0.820 & 0.711 & 0.507 & 1.364\\
    @Ours & \textbf{0.842} & \textbf{0.753} & \textbf{0.566} & \textbf{1.358} \\
    \bottomrule
  \end{tabular}
   
\end{table}

\subsubsection{RCA Strategy}
We evaluate three distribution alignment methods: MMD \cite{rozantsev2018beyond}, COD \cite{yang2024cod}, and our Rank-weighted Conditional Alignment (RCA). As shown in Table~\ref{tab:6}, COD, which leverages quality priors for conditional alignment, outperforms the marginal MMD baseline, highlighting the benefit of quality-aware alignment based on scores. Compared with COD, our RCA further improves performance, boosting SROCC by over 5.9\%, demonstrating the effectiveness of incorporating misranking-aware weighting.

\begin{table}[H]
\caption{Ablation study on our RCA strategy.}
\label{tab:6}  
  \centering
  \small
  \begin{tabular}{lcccc}
    \toprule
    Methods & PLCC↑ & SROCC↑ & KROCC↑ & RMSE↓ \\
    \midrule
    MMD & 0.803 & 0.695 & 0.507 & 1.432\\
    COD & 0.820 & 0.711 & 0.517 & 1.396\\
    RCA & \textbf{0.842} & \textbf{0.753} & \textbf{0.566} & \textbf{1.358} \\
    \bottomrule
  \end{tabular}
\end{table}

\subsubsection{Dual-Domain Augmentation Module}
This section evaluates the generalization of the dual-domain augmentation module. As shown in Table~\ref{tab:7}, DA(W) and DA(S) denote the module's performance on the WPC \cite{su2019perceptual}  and SJTU-PCQA \cite{yang2020predicting} datasets, respectively. Although a performance drop is observed on the unseen WPC \cite{su2019perceptual} dataset, this is expected because WPC \cite{su2019perceptual}  exhibits distortion types and perceptual characteristics that differ markedly from other datasets, resulting in a larger domain gap. Nevertheless, the dual-domain approach still surpasses single-domain mixing, showing stronger robustness under severe cross-dataset shifts.

\begin{table}[H]
\caption{Ablation study on dual-domain augmentation module.}
\label{tab:7}
  \centering
  \small
  \begin{tabular}{lcccc}
    \toprule
    Methods & PLCC↑ & SROCC↑ & KROCC↑ & RMSE↓ \\
    \midrule
    W/o DA(W) & 0.257 & 0.281 & 0.189 & 21.961\\
    Ours(W) & 0.303 & 0.296 & 0.201 & 21.959\\
    W/o DA(S) & 0.834 & 0.726 & 0.543 & 1.364 \\
    Ours(S) & \textbf{0.842} & \textbf{0.753} & \textbf{0.566} & \textbf{1.358}\\
    \bottomrule
  \end{tabular}
\end{table}

%% file: sec/5_con.tex
\section{Conclusion}
\label{sec:con}
In this paper, we propose a novel domain adaptation quality assessment framework named QD-PCQA, which leverages prior quality knowledge from images to predict point cloud quality. This framework consists of two key components: a Rank-weighted Conditional Alignment (RCA) strategy that adaptively aligns features based on quality-level consistency, and a Quality-guided Feature Augmentation (QFA) strategy that performs hierarchical, dual-domain, quality-aware style mixup. These components jointly preserve perceptual quality sensitivity and enhance generalization under domain shifts. Overall, QD-PCQA provides new insights into intrinsic correlations between different media and consistently delivers robust, competitive results.

%% file: sec/6_thank.tex
\section*{Acknowledgments}
This work is supported by the National Natural Science Foundation of China (62372036, 62120106009, U22A2022, 62332017), and the Ministry of Education, Singapore (Tier 1 RG103/24).